\pdfoutput=1

\documentclass[11pt]{article}

\usepackage{acl}

\usepackage{times}
\usepackage{latexsym}

\usepackage{graphicx} 
\graphicspath{ {./Images/} }

\usepackage{xcolor}
\usepackage{comment}
\usepackage{hyperref}
\usepackage{amsmath}
\usepackage{amssymb}
\usepackage{makecell}
\usepackage{tabularray}
\usepackage{algorithm}
\usepackage{algpseudocode}
\usepackage{pifont} 
\usepackage[most]{tcolorbox}
\usepackage{multirow}

\definecolor{navyblue}{RGB}{213, 224, 244}
\definecolor{forestgreen}{RGB}{181,225,167}
\definecolor{dirt}{RGB}{253,219,110}
\definecolor{salmon}{RGB}{250, 219, 216}
\newtcbox{\highlight}[1]{%
    on line,
    boxsep=0pt,
    left=2pt,
    right=2pt,
    top=2pt,
    bottom=2pt,
    boxrule=0pt,
    arc=0pt,
    outer arc=0pt, 
    colback=#1!100 
}

\usepackage[T1]{fontenc}

\usepackage[utf8]{inputenc}

\usepackage{microtype}

\usepackage{inconsolata}

\title{Over-Reasoning and Redundant Calculation of Large Language Models}

\author{Cheng-Han Chiang \\
  National Taiwan University,\\ Taiwan\\
  \texttt{dcml0714@gmail.com} \\\And
   Hung-yi Lee \\
  National Taiwan University,\\ Taiwan \\
  \texttt{hungyilee@ntu.edu.tw} \\}

\begin{document}
\maketitle
\begin{abstract}
Large language models (LLMs) can solve problems step-by-step.
While this chain-of-thought (CoT) reasoning boosts LLMs' performance, it is unclear if LLMs \textit{know} when to use CoT and whether those CoT are always necessary to answer the question. 
This paper shows that LLMs tend to generate redundant calculations and reasoning on a manually constructed math QA dataset, \textbf{\textit{GSM8K-Zero}}.
GSM8K-Zero is constructed such that the questions can be answered without any calculations, but LLMs, including Llama-2 models and Claude-2, tend to generate lengthy and unnecessary calculations to answer the questions.
We also conduct experiments to explain why LLMs generate redundant calculations and reasonings.
GSM8K-Zero is publicly available at \url{https://github.com/d223302/Over-Reasoning-of-LLMs} and \url{https://huggingface.co/datasets/dcml0714/GSM8K-Zero}.

\end{abstract}

\section{Introduction}

LLMs have shown impressive performance on various tasks~\citep{NEURIPS2020_1457c0d6,ouyang2022training,touvron2023llama}, and CoT reasonings~\citep{wei2022chain, kojima2022large} further boost the LLMs' performance.
To make LLMs better learn to think step-by-step, it is common to enhance the CoT ability of LLMs by adding training samples that contain CoT reasoning in instruction tuning~\citep{longpre2023flan}.

While step-by-step reasoning is useful, it lengthens the LLM's output.
Longer outputs imply longer inference latency and cost more money on commercial LLMs.
Moreover, if LLMs give long answers to questions that can be easily solved without complex reasoning, this may confuse the users and make them falsely believe that the questions are difficult and must be solved step-by-step.
It is thus more desirable that LLM only triggers CoT reasoning ability when necessary, but it is unclear if LLMs know when to think step-by-step and only generate step-by-step reasoning when necessary.
Prior works on CoT mostly focus on the correctness of the CoT reasoning steps or whether the reasoning steps are faithful to the question and support the final answer~\citep{ye2022the,golovneva2023roscoe}.
However, no past study focuses on the redundancy in CoT reasonings.

\begin{figure}[t]

\centering
\includegraphics[clip, trim = 0px 5px 0px 0px,width=1.0\linewidth]{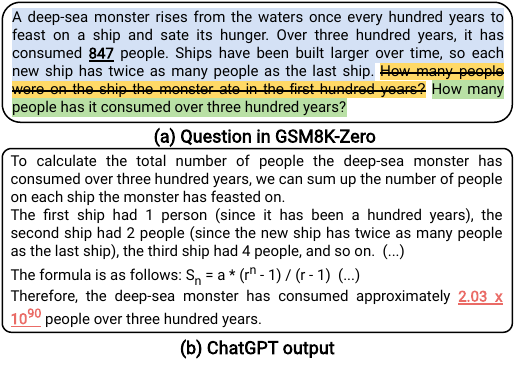}

\caption{ 
(a) A question in GSM8K-Zero.
The blue parts are the known information, and the orange part is the original question in GSM8K and is removed in GSM8K-Zero.
The green part is the new question in GSM8K-Zero.
(b) The answer generated by ChatGPT.
}
\label{fig:illustration.pdf}
\end{figure}

As an initiative to study the redundancy of LLM outputs, we aim to understand the following research question: Does LLM generate redundant reasonings when they clearly need not do so?
To study this question, we construct a math QA dataset, GSM8K-Zero, which contains trivial questions that can be answered without any calculations and reasoning.
Using this curated dataset, we can define the redundancy of output from LLMs.
We evaluate seven LLMs trained with reinforcement learning with human feedback (RLHF)~\citep{ouyang2022training}, and we find that LLMs tend to generate redundant calculations that complicate the responses and sometimes lead to the wrong answer.
To explain our observation, we show that GPT-4~\citep{openai2023gpt4} and ChatGPT~\citep{ChatGPT}, which are widely used in gathering the preference data for training a reward model in RLHF~\citep{guo-etal-2023-hc3,gpt4all}, show a strong preference towards long answers that contain redundant calculations, even if the long answers are incorrect.

Our contributions are summarized as follows:
\begin{itemize}
    \item To the best of our knowledge, we are the first to study the redundancy of LLM outputs.
    \item We construct and release a dataset, GSM8K-Zero, which reveals the LLMs' tendency to generate redundant reasonings.
    \item We show that LLMs tend to generate redundant calculations on math questions that can be answered without any calculation.
    \item We show that LLMs' tendency to generate long answers may stem from the imperfect reward model that prefers longer answers regardless of their correctness.
\end{itemize}

\section{Dataset: GSM8K-Zero}

\subsection{Construction of GSM8K-Zero}
\label{subsection: Construction of GSM8K-Zero}
To study LLMs' tendency for redundant calculations, we created \textbf{GSM8K-Zero} from GSM8K~\citep{cobbe2021training}. 
A question in GSM8K comprises (1) the \highlight{navyblue}{\textbf{known} information} (blue parts in Figure~\ref{fig:illustration.pdf}) and (2) \highlight{dirt}{a query for an \textbf{un-}} \highlight{dirt}{\textbf{known}} quantity (orange parts in Figure~\ref{fig:illustration.pdf}). 
Using questions in GSM8K, we aim to create questions whose answers are directly stated in the questions and can be obtained without any calculations.

We use the following procedure to achieve this goal.
The following procedure is best read with Figure~\ref{fig:illustration.pdf}(a).
Given a question in GSM8K, we remove the last sentence from the question that queries for an \highlight{dirt}{unknown variable} and keep the \highlight{navyblue}{known information}.
Next, we generate \highlight{forestgreen}{a} \highlight{forestgreen}{question that asks the value of a known variable} (green parts in Figure~\ref{fig:illustration.pdf}(a)) based on the \highlight{navyblue}{known} \highlight{navyblue}{information} and append the question behind the \highlight{navyblue}{known information}.
The question is generated by randomly selecting a number in the \highlight{navyblue}{known} \highlight{navyblue}{information} as the ground truth answer and using few-shot prompting to generate a question whose answer is the selected ground truth using ChatGPT.
We then use GPT-4 to answer the newly generated question. 
If GPT-4's answer deviates from the ground truth answer, the question is discarded.
We randomly select 3,500 questions from GSM8K's training set\footnote{In our preliminary experiment, we find that our results also hold when we use the testing set of GSM8K to construct the questions in GSM8K-Zero} and obtain 2,978 question-answer pairs after the above procedure.

Based on a manual inspection of 250 random question-answer pairs by the authors, we estimate that about 85\% of question-answer pairs in GSM8K-Zero are valid.
Refer to Appendix~\ref{appendix: Manual Review by the Authors} for a detailed description of our manual inspection of GSM8K-Zero.

\subsection{Evaluating Redundancy}
\label{subsection: Evaluating Redundancy}
We define redundant outputs as \textbf{any superfluous information in LLM responses that are not required for accurately answering the question}. 
Measuring this redundancy is often challenging for existing datasets.
However, GSM8K-Zero offers an easy way to evaluate LLM output's redundancy due to its unique nature: questions can be answered without any calculations since the answers are explicitly stated within the questions.
If an LLM's answer includes calculations, it is deemed redundant. 
We identify mathematical operators ($\times,+$, and $=$) in LLM outputs by a regular expression and say that the LLM's answer is redundant whenever mathematical operators are found.

\section{Experiments}
We test LLMs on GSM8K-Zero in zero-shot, as zero-shot inference closely mirrors most users' practical use of \textit{LLMs-as-assistants}. 
Instead of leveraging advanced prompting techniques like zero-shot CoT~\citep{kojima2022large} or Plan-and-Solve~\citep{wang-etal-2023-plan}, we present a single question to the LLM and take its response. 
For each question, we sample one response from the LLM.
In our preliminary experiments, we find the observations in our paper are robust toward the hyperparameters used for sampling outputs from LLMs.

Our evaluation encompasses proprietary LLMs, such as GPT-4, ChatGPT, Claude-2~\citep{claude2}, and PaLM (\texttt{text-bison-001})~\citep{anil2023palm}, and open-source ones like Llama-2-chat models of different sizes~\citep{touvron2023llama}.
We assess LLMs' performance on GSM8K-Zero using two metrics:
(1) \textbf{Redundancy}: Determined by the percentage of LLM answers containing numerical operators like $\times,+$, and $=$.
(2) \textbf{Accuracy}: Accuracy measures how often the LLM's answer, extracted using a regular expression, aligns with the GSM8K-Zero ground truth.

\begin{table}[t]
    \centering
    \begin{tabular}{c|c|ccc}
        \hline
         \multirow{ 2}{*}{\textbf{Models}}& \multirow{ 2}{*}{\textbf{\underline{Red.}}} &\multicolumn{3}{c}{\textbf{\underline{Accuracy}}}\\
          &    &Avg.& Cal. \ding{55} & Cal. \ding{51}\\
         \hline
         \multicolumn{5}{c}{\textit{Proprietary LLMs}} \\
         \hline
         GPT-4&   11.7 &100.0$^\dagger$ & 100.0$^\dagger$ & 100.0$^\dagger$\\
         ChatGPT&   47.1 &79.7&  96.6& 60.7\\
         Claude-2&   74.7&88.4& 98.8& 84.8\\
         PaLM &   29.2 &40.9&  40.9& 40.6\\
         \hline
         \multicolumn{5}{c}{\textit{Open-source LLMs (Llama-2)}} \\
         \hline
         70b-chat&   80.3 &54.5& 87.7& 46.3\\
         13b-chat&   88.3 &39.9&  86.0& 33.8\\
         7b-chat&   88.6 &41.4&  80.2& 36.3\\
         \hline
    \end{tabular}
    \caption{The redundancy (\textbf{Red.}) and accuracy of LLMs' responses.
    We report the average accuracy (\textbf{Avg.}) on all questions (second column), the accuracy for answers without calculation (Cal. \ding{55}, third column) and with calculation (Cal. \ding{51}, fourth column).
    $\dagger$: The accuracy of GPT-4 is 100\% by construction since we use GPT-4 to filter samples when constructing GSM8K-Zero.}
    \label{tab:main result}
\end{table}

\subsection{Main Results}
\label{subsection: Main Results}

We show the LLMs' performance on GSM8K-Zero in Table~\ref{tab:main result}.
First, we observe almost half of the LLMs we test have an accuracy lower than 50\% (second column in Table~\ref{tab:main result}).
Recall that the answers to the question in GSM8K-Zero can be easily extracted from the question without any calculations, which makes GSM8K-Zero more like an extractive QA than a math QA.
As simple as this dataset is, some LLMs still cannot perform well on it.

Next, we turn our attention to the redundancy in the answers.
It can be seen that both proprietary and open-source LLMs generate redundant calculations and reasoning to answer the questions.
ChatGPT yields unnecessary calculations in their answers in almost half of the answers, and all Llama-2 models generate lengthy reasoning steps and redundant calculations in more than $80\%$ of their responses while they are not explicitly prompted to do so.

We show some answers with redundancy generated by different LLMs in Table~\ref{tab:output example}.
By inspecting the outputs from LLMs, we find that in most cases, LLMs solve all the unknown variables in the questions, which are not asked in the questions (\#2 in Table~\ref{tab:output example}).
This behavior is problematic for a helpful assistant since it complicates the responses and may falsely lead the users to think solving all the unknown variables is necessary to arrive at the final answer.
We also find that the LLMs sometimes only provide the values of the unknown variables but do not answer the value asked in the question (\#1 in Table~\ref{tab:output example}), showing that LLMs cannot follow user instructions very well in these cases.

\begin{table}[t]
    \centering
    \begin{tabular}{c|cc}
        \hline
         \textbf{Model} & \textbf{Redundancy}  &  \textbf{Accuracy}\\
         \hline
         ChatGPT& 25.7&  83.6\\
         Claude-2& 40.7&  88.5\\
         \hline
         Llama-2-70b-chat& 54.4&  73.3\\
         Llama-2-13b-chat& 45.8&  65.5\\
         Llama-2-7b-chat& 32.7&  68.3\\
         \hline
    \end{tabular}
    \caption{The redundancy and accuracy of answers from LLMs when allowing LLMs not to use CoT.}
    \label{tab:no CoT}
\end{table}

\begin{table*}[t]
\footnotesize
    \centering
    \begin{tblr}{
      colspec = {X[m, 0.7em] X[j, 5.3em] X[j, 12em] X[j, 28em]}, 
      row{1} = {font=\bfseries}
    }
        \hline
        ID & Type & Question &  LLM Output \\
        \hline
        \#1 & LLM solves the unknown variables but does not answer the question &
        Albert is wondering how much pizza he can eat in one day. He buys 2 large pizzas and 2 small pizzas. A large pizza has \underline{\textbf{16}} slices and a small pizza has 8 slices. How many slices does a large pizza have? &
        (\texttt{Llama-2-70b-chat}) If Albert buys 2 large pizzas and 2 small pizzas, and a large pizza has 16 slices, then we can calculate the total number of slices of pizza that Albert can eat in one day as follows: 2 x 16 = 32 slices from the large pizzas + 2 x 8 = 16 slices from the small pizzas = Total slices of pizza that Albert can eat in one day = 48 slices. Therefore, Albert can eat 48 slices of pizza in one day. \\
        \hline
        \#2 & LLM solves the unknown variables and yields the correct answer&
        A deep-sea monster rises from the waters once every hundred years to feast on a ship and sate its hunger. Over three hundred years, it has consumed \underline{\textbf{847}} people. Ships have been built larger over time, so each new ship has twice as many people as the last ship. How many people has it consumed over three hundred years?&
        (\texttt{gpt-4}) The deep-sea monster consumes ships every 100 years, and over 300 years it would have consumed 3 ships. Given that each new ship has twice as many people as the last ship, it means there's a series of doubling. Let's assume the number of people on the first ship is X. Then the number of people on the second ship (which is twice the first) would be 2X and the number of people on the third ship (twice the second ship) would be 4X. Given that the total number of people consumed over three centuries is 847, we can write the equation: X + 2X + 4X = 847 This simplifies to 7X = 847  To find X (the number of people on the first ship), we divide 847 by 7, so X = 121. So, over three hundred years, the Monster has consumed 121 people on the first ship, 242 people on the second ship, and 484 people on the third ship. The total is 847 people.
        \\
        \hline
        \#3 & LLM generates incorrect calculations and yields an incorrect answer &
        John is a hack author. He writes a book every 2 months. He has been writing for \underline{\textbf{20}} years and has earned an average of \$30,000 per book. How many years has he been writing? &
        (\texttt{Llama-2-13b-chat}) To find out how many years John has been writing, we need to divide the total number of books he has written by the number of books he writes per year. Let's say John has written x books in total. Since he writes a book every 2 months, he has written: \highlight{salmon}{x = 20 years $*$ 2 months/year = 40 books}. We know that John has earned an average of \$30,0000 per book, so the total amount he has earned is: Total income = 40 books * \$30,000 = \$1,200,000 We can now solve for x: \$1,200,000 = 20 years $*$ 2 months/year x = \$1,20,000 / \$20,000. x = 10 years. Therefore, John has been writing for 10 years.\\
        \hline
        \#4 & LLM hallucinates variables that are not stated in the question and yields a wrong answer &
        During her first year playing soccer, Tara's dad attended \underline{\textbf{90\%}} of the games she played. In her second year playing soccer, Tara's dad attended 4 fewer games than he did in the previous year. What percentage of the games did Tara's dad attend during her first year playing soccer?&
        (\texttt{Llama-2-13b-chat}) During her first year playing soccer, Tara's dad attended 90\% of the games she played, which means he attended 90/100 = 9 games \highlight{salmon}{out of 10}. In her second year playing soccer, Tara's dad attended 90\% of the games, but instead of attending 9 games, he attended 90\% of \highlight{salmon}{10 games} minus 4 games, which is: 90\% x \highlight{salmon}{10} - 4 = 90\% x 6 = 54\% or 6 games. So, Tara's dad attended 54\% = 54/100 = 5 games out of \highlight{salmon}{the total of 10 games} played by Tara during her first year playing soccer. Answer: 5 games or 50\% of the games during Tara's first year playing soccer.\\
        \hline
    \end{tblr}
    \caption{Example of different types of redundant calculations from various LLMs.
    The ground truth answer to the question is \underline{\textbf{underlined}} in the question.
    We replace the newline character from the LLMs' output with space or period for easier illustration.
    We \highlight{salmon}{highlight} the wrong calculations and hallucinated variables.}
    \label{tab:output example}
\end{table*}

After discussing redundancy and accuracy independently, we want to know if redundant calculation co-occurs more often with wrong answers.
We separate the model outputs into two groups: one that contains calculations and another that does not have calculations, and we calculate the accuracy for the two groups.
The results are shown in the two rightmost columns in Table~\ref{tab:main result}.
When the LLM's answers contain calculations, the accuracy drops significantly for almost all models except for PaLM.
By randomly browsing the wrong answers that include calculations of models except PaLM, we find that sometimes LLMs hallucinate variables not specified in the questions (\#4 in Table~\ref{tab:output example}).
Sometimes, LLMs make calculation errors, leading to the wrong answer (\#3 in Table~\ref{tab:output example}).
This shows that redundant calculations not only waste time and resources but can also hurt the LLM's performance due to calculation errors and incorrect reasoning.

For the outputs of PaLM that contain calculations, we observe that PaLM often first generates an Arabic number as the answer, followed by some calculations as the explanation.
In this case, the numeric answer of PaLM does not depend on the calculations, so even if the calculation and reasoning following the answers are wrong, they cannot affect the answer.
This makes the accuracy of answers with and without calculation similar in the case of PaLM.

\subsection{Do LLMs Know When to Use CoT?}
Section~\ref{subsection: Main Results} reveals that LLMs can generate redundant calculations and unnecessary CoT reasoning steps.
This is possibly because, during instruction tuning, LLMs are trained to generate CoT reasoning for mathematical problems \textbf{when the input instruction does not specify how to solve the question}, forcing them to apply CoT on every question that \textit{looks like} a mathematical question.
Hence, we are curious whether LLMs can drop the CoT reasoning and calculations \textbf{when properly instructed}.
To explore this possibility, we append the following instruction after the questions in GSM8K-Zero: "\texttt{If the question is simple enough, you can omit the step-by-step reasoning and just give the answer.}"
Here, we only test on the LLMs that generate answers with higher redundancy in Section~\ref{subsection: Main Results}.

The results are shown in Table~\ref{tab:no CoT}.
We can see that when LLMs are allowed to omit step-by-step reasoning, the redundancy of the LLMs significantly drops compared with Table~\ref{tab:main result} while the accuracy significantly boosts for almost all models.
The decrease in output redundancy implies that LLMs do know that some questions in GSM8K-Zero are easy enough to answer directly.
However, even when they are allowed to omit step-by-step reasoning, the redundancy in these LLMs is still higher than 25\%.
This means that LLMs cannot always correctly infer the difficulty and whether step-by-step reasonings are necessary for the questions.

%

\section{Why Do LLMs Generate Redundant Calculations?}
\label{section: Why Does LLMs Generate Redundant Calculations?}

After seeing that LLMs produce excessive calculations, we seek to understand why. 
We speculate that the reward models (RMs) in RLHF might favor more verbose outputs over concise ones, making RLHF-trained models prone to generate lengthy output even if it is redundant. 
To test this hypothesis, we would like to compare RM's preference between long and short answers. 
However, we cannot access RMs used to train ChatGPT or Llama models.
As a workaround, we use ChatGPT and GPT-4 as the proxy of the RMs; we call these models \textit{proxy RMs} in this case.
To obtain the preference of the proxy RMs, we give proxy RMs some instructions, a question in GSM8K-Zero, a pair of long and short answers, and ask the model to select a better answer.
We follow the instructions used in~\citet{zheng2023judging}, which asks the proxy RMs to consider the \textbf{accuracy} and \textbf{helpfulness} of the answer.
The experiment is repeated by inverting the order of the short and long answers to counteract potential position bias. 
Using ChatGPT or GPT-4 as the proxy RMs is reasonable, as these models should learn the preferences of their RMs during RLHF.
Additionally, prior works have used ChatGPT and GPT-4 to generate the preference data to train the RMs~\citep{gpt4all}, so the preference of ChatGPT or GPT-4 can reflect the preference of RMs.

We prepare the long and short answers as follows: 
To collect long answers, we collect ChatGPT's answers to questions in GSM8K-Zero, select those with redundant calculations, and group those answers into two: correct answers and incorrect answers, with approximately 100 samples in each group.
The 100 samples in the correct-answer group are reviewed by one of the authors to ensure that the answer is correct instead of a false positive due to imperfect regular expressions when extracting the answer from the LLM's response.
The same procedure is done for the 100 samples in the incorrect-answer group.
Next, for each long answer collected, we construct a short answer counterpart by the template, "\texttt{The answer is [[ground truth]]}", where \texttt{"[[ground truth]]"} is filled in with the ground truth in GSM8K-Zero.

\begin{figure}[t]

\centering
\includegraphics[clip, trim = 20px 18px 20px 18px,width=1.0\linewidth]{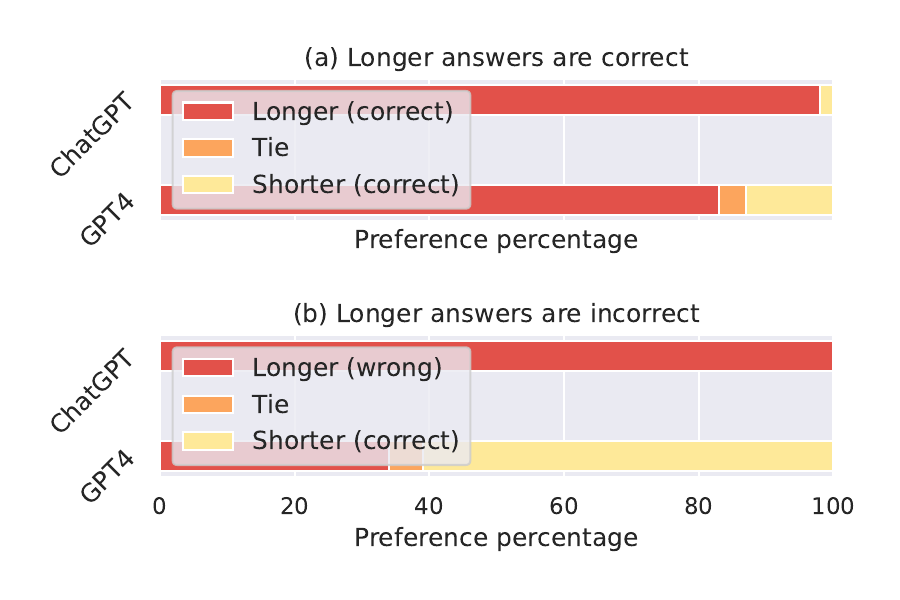}

\caption{ 
The preference of GPT-4 and ChatGPT between longer and shorter answers.
(a) The case when the longer answers are correct.
(b) The case when the longer answers are incorrect.
}
\label{fig:judge.pdf}
\end{figure}

The preference of proxy RMs between long and short answers is shown in Figure~\ref{fig:judge.pdf}.
First, we observe that when both the long and short answers are correct (Figure~\ref{fig:judge.pdf}(a)), both GPT-4 and ChatGPT prefer long answers.
By scrutinizing the evaluation results, we find that GPT-4 and ChatGPT frequently complain about the shorter answer to "only answer the question without any further details," while the long answer "shows more information."
However, when reading the long answers, the authors find it hard to locate the answer to the question since the model outputs too much unnecessary information and complicates the problem, making the answer unhelpful.
Next, when the long answer is incorrect and the short answer is correct (Figure~\ref{fig:judge.pdf}(b)), we find that ChatGPT consistently prefers lengthy but wrong answers.
While GPT-4 successfully prefers the short and correct answer in 61\% of the cases, GPT-4 still votes for long but wrong answers in 34\% of the cases.
Overall, the results in Figure~\ref{fig:judge.pdf} show that proxy RMs strongly prefer long outputs that contain redundant calculations and unnecessary reasoning, even if the final answer is wrong!
If we use the proxy RMs' preference data collected in this section, it is easy to think that we will obtain RMs that favor lengthy output, eventually leading to an LLM that generates redundant calculations.
We repeat the above experiment using the answers from Llama-7b-chat and observe a similar result.

\section{Conclusion}
In this paper, we construct GSM8K-Zero to illustrate the redundancy in the output from LLMs.
We show that LLMs tend to generate redundant calculations and unnecessary reasoning, sometimes leading to a wrong answer.
We reveal that LLMs may not differentiate questions requiring step-by-step reasoning from simpler ones, suggesting a possible direction for improving LLMs.
To explain our observation, we use proxy RMs and find that these models prefer lengthy answers even if they are wrong.
Through this paper, we hope future researchers can focus more on the redundancy of the outputs of LLMs and develop training techniques to teach LLMs when to think step-by-step.

\section*{Limitations}
The main limitation of our paper is that we only study redundancy on a manually constructed dataset, GSM8K-Zero.
The reason is that it is easier to define and calculate redundancy on GSM8K-Zero; we believe this is an ample contribution since it is a phenomenon never mentioned in the literature.
While exploring redundancy on other existing datasets will be interesting, we leave it to future works.

Another limitation of our paper is that we rely on ChatGPT and GPT-4 to construct GSM8K-Zero, so noises in the constructed dataset are inevitable.
We emphasize that future researchers need to keep the noises in the dataset in mind and take special caution when interpreting the results evaluated on GSM8K-Zero.
To understand the noises in the dataset, the authors randomly selected 250 samples from GSM8K-Zero and reviewed them.
As stated in Section~\ref{subsection: Construction of GSM8K-Zero}, we estimate that 85\% of question-answer pairs in GSM8K-Zero are valid.
We present the details about our manual review of the dataset in Appendix~\ref{appendix: Manual Review by the Authors}.
We also discuss that our results and observations in the main content still hold when considering the noises in the dataset.

Last, since our paper is a short paper, an obvious limitation is that there is still a lot to explore, but we cannot include them in our paper.
While we deem our paper's main content to be self-contained, we include some potential questions that might be raised by curious and enthusiastic readers in Appendix~\ref{appendix: FAQ} (FAQs section).

\section*{Ethical Statements}
We do not see our work to have possible harmful outcomes.
We follow the ACL ethical guidelines when conducting the research in this paper.

\section*{Acknowledgements}
We want to thank the reviewers for providing detailed feedback and actionable suggestions, which helped us strengthen our paper.
We also want to thank the senior committee members for monitoring the reviewing process.
Cheng-Han Chiang is supported by a Google PhD Fellowship and a Ph.D. scholarship program by Delta Electronics.
We thank the National Center for High-performance Computing (NCHC) of National Applied Research Laboratories (NARLabs) in Taiwan for providing computational and storage resources.

\bibliography{anthology,custom}

\appendix

\section{FAQs}
\label{appendix: FAQ}
\begin{itemize}
    \item [Q1] This paper only studies RLFH models. 
    What about LLMs that are not RLHF-trained? 
    Do they also show redundancy in their outputs?
    \item [A1] Yes, non-RLHF-trained LLMs also show redundancy in their outputs on GSM8K-Zero.
    We use Alpaca~\citep{alpaca} and Vicuna~\citep{vicuna2023} and find them to also generate redundant outputs in 40\% of the cases. 
    We do not report the results in the main paper since the outputs from Alpaca and Vicuna are quite messy, and it is hard to calculate the accuracy using regular expressions.
    \item [Q2] In Section~\ref{section: Why Does LLMs Generate Redundant Calculations?}, is it possible that the wrong and long answers generated by ChatGPT are correct, making the proxy RMs prefer those long answers?
    For example, when using regular expressions to calculate accuracy, there might be some cases that regular expressions cannot handle.
    \item [A2] This is highly unlikely to happen.
    This is because one of the authors manually reviews the long answers (100 correct and 100 wrong ones) used in Section~\ref{section: Why Does LLMs Generate Redundant Calculations?}.
    Thus, the wrong answers are assured to be wrong, and the correct answers are assured to be correct.
    Since the authors cannot review all the answers that contain calculations, we only randomly sample approximately 100 correct and 100 wrong answers with calculations and include them in the results in Figure~\ref{fig:judge.pdf}.
\end{itemize}

\section{More Information about GSM8K-Zero}
\subsection{Dataset Cards}
GSM8K-Zero is constructed from GSM8K~\citep{cobbe2021training}. 
Since GSM8K does not include the dataset license, we are unsure what license to release GSM8K-Zero.

\subsection{Manual Review by the Authors}
\label{appendix: Manual Review by the Authors}
The authors randomly sample 250 samples from GSM8K-Zero to understand the quality of the samples and whether using regular expression to calculate accuracy has a high precision.
The human (author) evaluation is conducted in the following steps:
First, we randomly sample 125 samples from the answers of ChatGPT that are correct together with their corresponding questions, and we sample 125 samples for the answers of ChatGPT that are incorrect together with their corresponding questions.
Recall that the accuracy is calculated using regular expressions.
We search for the first or last number that appears in the last sentence of the model's response, and we count the model response to be accurate if the ground truth matches the number extracted by regular expressions.
While this process may falsely consider the model to be correct when the model's answer is wrong, we find that this merely happens during our manual review of 250 answers from ChatGPT.
We separately sample questions that ChatGPT correctly answered and questions that ChatGPT got wrong because those two groups of questions might be systematically different.

Given a question, an answer from ChatGPT, and the ground truth answer, one of the authors labels the sample into four categories:
\begin{enumerate}
    \item  The ground truth is correct, and the answer from ChatGPT is correct
    \item The ground truth is wrong, while the answer from ChatGPT is correct (matches the real ground truth)
    \item  The ground truth is correct, but the answer from ChatGPT is wrong
    \item The question is invalid, including that ground truth is wrong, the question cannot be answered without calculation, or the question is ambiguous.
\end{enumerate}

We find that for questions that ChatGPT is correct, 89\% of questions are valid, and the ground truth answer is always correct.
However, we find that for 7\% of the questions, ChatGPT's answer is wrong, but we count it as correct due to imperfect parsing of regular expressions.
For questions that ChatGPT is inaccurate, about 70\% of the questions are valid, and the ground truth is wrong in 2\% of the cases. 
Only in 4\% of the cases, the regular expression we use considers the answer of ChatGPT to be wrong when it is correct.
Considering that ChatGPT's accuracy is about 80\%, we estimate that the proportion of invalid questions in GSM8K-Zero is 14.8\%.

\subsubsection{Does Invalid Questions Affect the Results?}
Readers may be concerned about whether the invalid questions change the observations in the main content.
The short answer is no.
We explain as follows:
For the redundancy shown in Table~\ref{tab:main result}, if the model generates CoT reasonings and calculations for those invalid questions, then the redundancy should be around 15\%.
But clearly, all model except GPT-4 has a redundancy much higher than 15\%.
Thus, LLMs can still generate a lot of redundant calculations for other valid questions.
As for GPT-4, we still find that it does generate redundant calculations in some cases.
\#2 in Table~\ref{tab:output example} is such a case.

Next, for accuracy, even if LLMs are wrong for all the invalid questions, their accuracy should be around 85\% if they get all the valid questions correct.
However, this is clearly not the case for all LLMs except Claude-2.
Next, for the rightmost column in Table~\ref{tab:main result}, if we assume that all the invalid samples happen to be the samples that LLMs include calculations in the answer, the accuracy in this column should increase.
However, by some simple maths, the readers can easily verify that even considering this, the accuracy of answers containing calculations is still much lower than that of answers that do not include calculations.
Thus, our observation in the paper still holds.

\section{Prompts}
We list the prompts we use in this section.

\paragraph{Prompts for question generation using ChatGPT in Section~\ref{subsection: Construction of GSM8K-Zero}}

\textbf{System prompt}: \texttt{You are a helpful assistant. You need to answer the questions of the user accurately. 
You need to strictly follow the instructions.}

\textbf{User prompt}

\texttt{
Your task is to convert a declarative sentence into a question and the answer to that question should be a number. Importantly, the answer (number) to the question should already be included in the original sentence. If the answer need to be obtained by calculation, the question is invalid. Even simple calculation is not allowed.  Keep the question as simple as possible. For example:\\
Example 1:\\
Original sentence: Alyssa, Keely, and Kendall ordered 100 chicken nuggets from a fast-food restaurant.\\
Answer (number only): 100\\
Question: How many chicken nuggets did Alyssa, Keely, and Kendall order?\\
Explanation: The number 100 already appeared in the original sentence, so the question fulfill the requirements.
}

\texttt{
Example 2: \\
Original sentence: Lilah's family gallery has 400 photos. \\
Answer (number only): 400 \\
Question: How many photos are there in Lilah's family gallery? \\
Explanation: The number 400 already appeared in the original sentence, so the question fulfill the requirements.
}

\texttt{
Example 3: \\
Original sentence: \{KNOWN\_INFO\} \\
Answer (number only): \{ANS\} \\
Question: 
}

The \texttt{\{KNOWN\_INFO\}} should be filled in with the \highlight{navyblue}{known information} in the original question, and the \texttt{\{ANS\}} should be filled in with the ground truth answer.

\paragraph{Prompts for using ChatGPT and GPT-4 as the proxy in Section~\ref{section: Why Does LLMs Generate Redundant Calculations?}}

\paragraph{System prompt}
\texttt{Please act as an impartial judge and evaluate the quality of the responses provided by two AI assistants to the user question displayed below. 
You should choose the assistant that follows the user's instructions and answers the user's question better. 
Your evaluation should consider factors such as the helpfulness, relevance, accuracy, depth, creativity, and level of detail of their responses. 
Begin your evaluation by comparing the two responses and provide a short explanation. 
Avoid any position biases and ensure that the order in which the responses were presented does not influence your decision. 
Do not allow the length of the responses to influence your evaluation. 
Do not favor certain names of the assistants. Be as objective as possible. 
After providing your explanation, output your final verdict by strictly following this format: "[[A]]" if assistant A is better, "[[B]]" if assistant B is better, and "[[C]]" for a tie.
}

\paragraph{User Prompt}

\texttt{[User Question]
\newline
\{question\}
\newline}
\texttt{
[The Start of Assistant A's Answer]
\newline
\{answer\_a\} 
\newline
[The End of Assistant A's Answer]}
\newline
\newline
\texttt{
[The Start of Assistant B's Answer]
\newline
\{answer\_b\}
\newline
[The End of Assistant B's Answer]}

\section{Sampling parameters of LLMs}
When using LLMs to generate the answer to questions in GSM8K-Zero, we set the temperature to 0.7 and keep all the other parameters as default.
We use Huggingface Transformers to run Llama-2.
\end{document}